\documentclass{ifacconf}

\usepackage{graphicx}      % include this line if your document contains figures
\usepackage{natbib}        % required for bibliography
\usepackage{amsmath}
\usepackage{algorithm}
\usepackage[noend]{algpseudocode} % 推荐这个，比老的 algorithmic 好用
\usepackage{amsfonts}
\usepackage{amssymb}
\usepackage{wrapfig}
\usepackage{graphicx}  
\usepackage{xcolor}
\usepackage[table]{xcolor}

\begin{document}
\begin{frontmatter}

% \title{Bridging Discrete Planning and Continuous Execution for Redundant Robot Manipulators } 
% % Title, preferably not more than 10 words.
% \thanks[footnoteinfo]{This work was supported in part by Youth S$\&$T Talent Support Programme of GDSTA (SKXRC2025468). }

% \author[First]{Teng Yan} 
% \author[First]{Yue Yu} 
% \author[First]{Yihan Liu} 
% \author[Second]{Bingzhuo Zhong}

% \address[First]{The Thrust of Artificial Intelligence, Information Hub, Hong Kong University of Science and Technology (Guangzhou), Guangzhou 511400, China (e-mail: tyan497@connect.hkust-gz.edu.cn, yyu704@connect.hkust-gz.edu.cn, yliu135@connect.hkust-gz.edu.cn).}
% \address[Second]{The Thrust of Intelligent Transportation, System Hub, Hong Kong University of Science and Technology (Guangzhou), Guangzhou 511400, China (e-mail: bingzhuoz@hkust-gz.edu.cn)}

\title{Bridging Discrete Planning and Continuous Execution for Redundant Robot Manipulators\thanksref{footnoteinfo}}

\thanks[footnoteinfo]{This work is supported by the Guangdong Basic and Applied Basic Research Foundation (No: 2026A1515010222), the Guangdong Provincial Project (No: 2024QN11X053), and the Youth S\&T Talent Support Programme of GDSTA (No: SKXRC2025468). Corresponding author: Bingzhuo Zhong.}

\author[AI]{Teng Yan}
\author[AI]{Yue Yu}
\author[AI]{Yihan Liu}
\author[AI]{Bingzhuo Zhong}

\address[AI]{Thrust of Artificial Intelligence, The Hong Kong University of Science and Technology (Guangzhou), Guangzhou 511400, China\\
(e-mail: \{tyan497, yyu704, yliu135\}@connect.hkust-gz.edu.cn; bingzhuoz@hkust-gz.edu.cn)}

% \author[First]{Teng Yan},
% \author[First]{Yue Yu},
% \author[First]{Yihan Liu},
% \author[First,Second]{Bingzhuo Zhong} 

% \address[First]{The Thrust of Artificial Intelligence, Information Hub, Hong Kong
% University of Science and Technology (Guangzhou), Guangzhou
% 511400, China (e-mail: tyan497@connect.hkust-gz.edu.cn, yyu704@connect.hkust-gz.edu.cn, yliu135@connect.hkust-gz.edu.cn).}
% \address[Second]{The Thrust of Intelligent Transportation, System Hub, Hong Kong
% University of Science and Technology (Guangzhou), Guangzhou
% 511400, China (e-mail: bingzhuoz@hkust-gz.edu.cn)}

\begin{abstract}
% Voxel-grid reinforcement learning is widely adopted for path planning in redundant manipulators due to its simplicity and reproducibility. However, direct execution through point-wise numerical inverse kinematics on 7-DoF arms often yields step-size jitter, abrupt joint transitions, and instability near singular configurations. This work proposes a bridging framework between discrete planning and continuous execution without modifying the discrete planner itself. On the planning side, step-normalized 26-neighbor Cartesian actions and a geometric tie-breaking mechanism are introduced to suppress unnecessary turns and eliminate step-size oscillations. On the execution side, a task-priority damped least-squares (TP-DLS) inverse kinematics layer is implemented. This layer treats end-effector position as a primary task, while posture and joint centering are handled as subordinate tasks projected into the null space, combined with trust-region clipping and joint velocity constraints. On a 7-DoF manipulator in random sparse, medium, and dense environments, this bridge raises planning success in dense scenes from about 0.58 to 1.00, shortens representative path length from roughly 1.53\,m to 1.10\,m, and while keeping end-effector error below 1\,mm—reduces peak joint accelerations by over an order of magnitude, substantially improving the continuous execution quality of voxel-based RL paths on redundant manipulators.

Grid-based reinforcement learning is widely used for path planning in redundant manipulators due to its simplicity and reproducibility. However, directly executing discretely planned voxel paths on a 7-DoF arm via point-wise numerical inverse kinematics (IK) often causes step-size jitter, abrupt joint transitions, and instability near singularities. This work addresses this common engineering issue by proposing an offline bridging framework that enables smooth continuous execution without modifying the discrete planner. On the planning side, step-normalized 26-neighbor Cartesian actions and a geometric tie-breaking rule reduce unnecessary turns and step oscillations. On the execution side, an existing task-priority damped least-squares (TP-DLS) IK method is adopted, where end-effector position is the primary task and posture regulation and joint centering are projected into the null space, with trust-region clipping and joint velocity limits. Experiments on a 7-DoF manipulator show that this bridge increases planning success in dense scenes from 0.58 to 1.00, shortens representative path length from 1.53\,m to 1.10\,m, and reduces peak joint accelerations by over an order of magnitude while keeping end-effector error below 1\,mm. These results demonstrate that grid-based RL paths can be made reliably executable through careful integration with established IK techniques.

\end{abstract}


\end{frontmatter}
%===============================================================================

\section{Introduction}

Redundant robot manipulators are widely used for collision-free motion while satisfying joint limits, avoiding singularities, and respecting actuator constraints. Classical sampling- and optimisation-based planners offer strong guarantees but require accurate models and careful tuning, making frequent layout or task changes costly. Reinforcement-learning (RL) planners on voxelised workspaces are therefore attractive: they learn a reusable value function on a finite state–action space and answer motion queries by table look-up. A common formulation discretises the task space into a voxel grid, defines a finite neighbourhood action set, and produces a discrete Cartesian path from start to goal.

When such a path is executed on a seven-degree-of-freedom (7-DoF) arm using pointwise numerical inverse kinematics (IK), planning and execution become misaligned. Each waypoint is solved independently, causing step-size fluctuations, abrupt joint transitions, and sensitivity to singularities and joint limits. This work addresses the widely observed engineering issue that discretely planned RL paths are difficult to execute continuously, by proposing an offline bridging framework that enables smooth execution without modifying the discrete planner.

\begin{figure}
\begin{center}
\includegraphics[width=8.4cm]{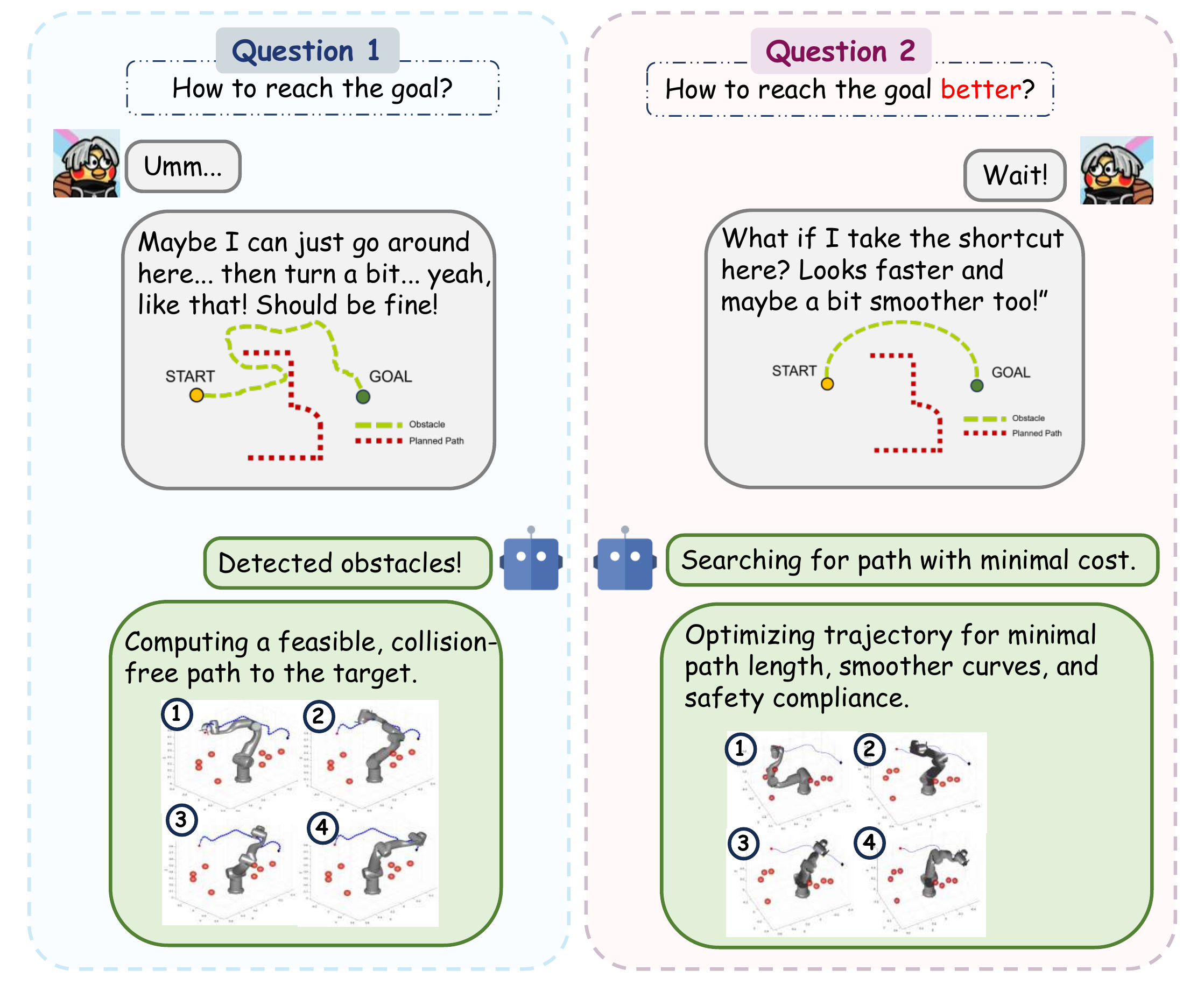}    
\caption{\textbf{Conceptual motivation}. Left: classical planning - feasible collision-free path (\textbf{Q1}); Right: practical deployment - high-quality path (\textbf{Q2}). The proposed method meets the second requirement by bridging a voxel RL planner with a continuous execution layer on a redundant manipulator.} 
\label{fig:intro}
\end{center}
\end{figure}

Figure~\ref{fig:intro} highlights this gap. For redundant manipulators, it is not enough to find a feasible path; the path should also be shorter, smoother, and compliant with safety and actuation constraints. This paper focuses on this aspect and adopts a task-priority damped least-squares (TP-DLS) inverse kinematics method as the execution layer.

A further difficulty stems from the geometry of the discrete action set. In a three-dimensional voxel grid, the standard 26-neighbourhood mixes axis-aligned, face-diagonal, and space-diagonal moves with different Euclidean lengths. Combined with $\varepsilon$-greedy exploration and approximate value functions, this anisotropy produces “saw-tooth’’ paths where long and short steps are interleaved and the end-effector oscillates around obstacles or meanders in unproductive directions. Paths become unnecessarily long, contain redundant turns, and visit homotopy classes that are hard to track smoothly in joint space.

Existing techniques address only parts of this interface. Path smoothing and time-parameterisation fit continuous trajectories through discrete waypoints and enforce velocity and acceleration bounds, but do not correct topological artefacts such as detours or oscillations. Damped least-squares (DLS) and task-priority IK improve robustness near singularities and support secondary goals such as joint-limit avoidance, yet are usually applied as pointwise solvers and assume already smooth references. As a result, the interface between discrete planning and continuous execution is often treated as a trivial concatenation rather than a design problem.

This paper treats that interface explicitly and proposes a lightweight bridge between voxel-based RL planners and continuous execution on redundant manipulators. The RL planner is used as a black box; only the geometry of its action space and the execution layer are modified. On the planning side, the 26-neighbourhood is replaced by a step-normalized Cartesian action set with a geometric tie-breaking rule that prefers small turning angles and larger estimated clearance, regularising path geometry without changing the RL update rule. On the execution side, a TP-DLS layer tracks a globally smoothed and locally subdivided end-effector path, using joint-centering in the null space, along with norm and per-joint increment limits, to enforce trust-region-like bounds on joint updates.

The proposed bridge is evaluated on a 7-DoF manipulator in MATLAB across random environments with sparse, medium, and dense obstacle distributions. Metrics at both planning and execution levels quantify path length, turning behaviour, clearance, joint-space smoothness, and dynamic feasibility. Results show that modest modifications at the planning and execution interfaces substantially improve planning success, shorten paths, and turn numerically feasible but physically unrealistic IK sequences into dynamically reasonable motions. The main contributions of this paper are:

\textbf{Mismatch analysis.} Identifies the structural mismatch between voxel-based RL paths and continuous execution on redundant manipulators, clarifying the roles of anisotropic action metrics and pointwise IK in generating artefacts.

\textbf{Bridge architecture.} Proposes a lightweight bridge that combines step-normalized 26-neighbour actions with geometric tie-breaking on the planning side and a TP-DLS execution layer with joint-centering and bounded joint updates.

\textbf{Benchmark and evidence.} Develops a benchmark suite and metric set for a 7-DoF arm and demonstrates improved path length, joint-space smoothness, and physical executability in cluttered environments.

\section{Related Work}
\subsection{RL-Based Motion Planning and Voxel Policies}
Recent surveys provide comprehensive overviews of deep reinforcement learning (DRL) in robotic-manipulator motion planning and identify persistent limitations in safety, sample efficiency, and execution robustness on high-DOF systems \citep{elguea2024review,han2023survey,zhang2026state}. For redundant manipulators, DRL-based frameworks have been shown to directly generate task-space motions while considering kinematic limits \citep{li2021general}. Research leveraging voxel or grid representations further demonstrates the suitability of discretized workspaces and finite action sets for RL-based planning \citep{landgraf2021reinforcement,james2022coarse}. The reliability of such discretized policies often hinges on the fidelity of the underlying 3D scene understanding. Recent advances in robust point cloud semantic segmentation and sparse object detection via diffusion models \citep{qu2025end,qu2026robust} provide potential pathways for enhancing the environmental awareness required for safe navigation in cluttered spaces.

A representative example in this direction is the use of RL for autonomous obstacle-avoidance path planning on six-axis robotic arms, where voxelized environments enable stable policy learning and reliable deployment \citep{jia2020path}.

Although these works validate the effectiveness of voxel and discrete-action policies, they typically focus on producing discrete end-effector paths and do not examine how such paths should be executed on redundant manipulators governed by continuous-time dynamics.

\subsection{Redundancy Resolution and Task-Priority Control}
For execution on redundant arms, damped least-squares (DLS) inverse kinematics has long been employed to bound joint velocities and mitigate singularity-induced instabilities \citep{deo1995overview}. Task-priority control organizes multiple objectives—such as end-effector tracking, posture optimization, and safety margins—into a hierarchical structure, projecting secondary tasks into the Jacobian null space \citep{nakamura1987task,basso2020task}. These classical formulations form the mathematical basis of the task-priority DLS (TP-DLS) execution layer used in this work.

In parallel, safety-focused architectures have been proposed to encapsulate learning-based controllers within verifiable supervisory layers \citep{zhong2021safe,zhong2023secure}. The present work does not introduce a new safety filter; instead, it shows that a suitably regularized TP-DLS layer, together with mild action-space constraints, already yields improved smoothness and safety when executing voxel RL policies on 7-DOF arms.

% \subsection{Trajectory Smoothing and Optimization}
% Beyond kinematic execution, substantial research has focused on smoothing piecewise-linear paths generated by sampling-based planners. Collision-aware spline constructions and time-parameterization approaches have been shown to reduce curvature, acceleration peaks, and discontinuities in task-space trajectories \citep{pan2012collision,kim2020task}. Optimization-based methods further integrate collision avoidance and dynamic feasibility into unified formulations \citep{liu2021creating}. These techniques, however, typically require solving a dedicated optimization problem for each path.

% In contrast, the present work introduces a lightweight bridging mechanism. Rather than invoking a full trajectory optimizer, it regularizes the geometry of the discrete planner and the differential update law of the execution layer, enabling existing voxel planners to be executed directly on redundant manipulators. From the RL perspective, sequence-modeling approaches that incorporate Q-value regularization—such as Q-Value Regularized Decision ConvFormer (QDC)—highlight the benefits of stabilizing return consistency during trajectory stitching \citep{ruan2024q}. This viewpoint motivates interpreting voxel policies as discrete sequences that require tailored bridging strategies at the geometric and control levels to interface with continuous robot dynamics.

Prior work has addressed the smoothing of piecewise-linear paths from sampling-based planners using collision-aware splines, time-parameterization, and optimization-based formulations to reduce curvature, acceleration peaks, and discontinuities in task space \citep{pan2012collision,kim2020task,liu2021creating}. These approaches, however, typically require solving a dedicated optimization problem for each path.

This work instead proposes a lightweight bridging mechanism. By regularizing the geometry of the discrete planner and the differential update law of the execution layer, voxel-based paths can be executed directly on redundant manipulators without invoking a full trajectory optimizer. From the RL perspective, sequence-modeling methods with Q-value regularization, such as QDC, emphasize stabilizing return consistency during trajectory stitching \citep{ruan2024q}, motivating the view that voxel policies form discrete sequences requiring dedicated geometric and control-level bridging to interface with continuous robot dynamics.

\section{Methodology: bridging discrete geometry and continuous physics}
The method is structured into two tightly coupled layers.

\begin{figure*}[htbp]
    \centering
    \includegraphics[width=\textwidth]{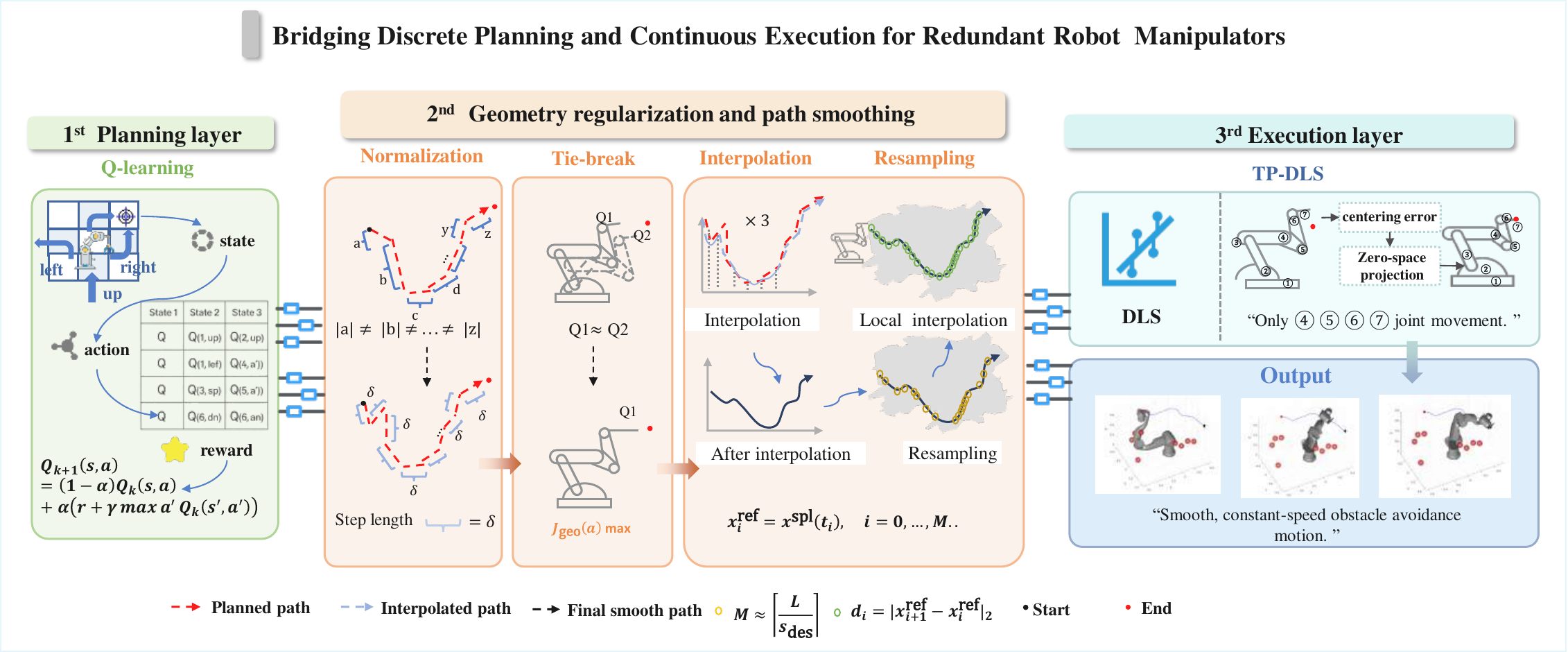}
    \caption{Overall architecture of the proposed bridging framework combining discrete Q-learning planning, geometric regularization, and continuous TP-DLS execution.}
    \label{fig:method}
\end{figure*}

\textbf{Geometry Layer.}
This layer operates in task space, where a voxel-based Q-learning planner generates a discrete end-effector path using normalized 26-connected actions together with a geometric tie-breaking mechanism.

\textbf{Physics Layer.}
This layer operates in joint space, where a task-priority damped least-squares (TP-DLS) inverse kinematics solver tracks a smoothed version of the planned path on a 7-DoF manipulator.

% The specific details for overall architecture can be found in \ref{fig:method}. And in the underlying implementation, the manipulator model internally stores joint configurations in row-vector format when interfacing with standard kinematic routines. For the theoretical development, however, joint configurations are denoted as column vectors $q \in \mathbb{R}^7$ to maintain consistency with conventional analytical notation. The difference between row and column representations is purely notational and does not affect the underlying kinematic definitions.

The overall architecture is shown in Fig.~\ref{fig:method}. In implementation, joint configurations are stored as row vectors for compatibility with standard kinematic routines, while theoretical analysis uses column vectors $q \in \mathbb{R}^7$ for consistency with conventional notation. This difference is purely notational and does not affect the underlying kinematics.

Let $p(q) \in \mathbb{R}^3$ denote the end-effector position obtained from the forward kinematic mapping. The corresponding geometric Jacobian is written as $J(q) \in \mathbb{R}^{6 \times 7}$. Its translational component, associated with the positional degrees of freedom, is denoted by $J_{p}(q) \in \mathbb{R}^{3 \times 7}$ and corresponds to the upper three rows of $J(q)$.

\subsection{Geometry layer: voxel-based Q-learning and action regularisation}

\subsubsection{Q-learning planner on a 3-D voxel grid}
The reachable workspace is discretised into a uniform three-dimensional grid of cubic voxels. Let $\mathcal{S} \subset \mathbb{R}^3$ denote the set of obstacle-free voxel centres. A tabular Q-learning agent is trained on this grid.

At each decision step, the agent occupies a state $s \in \mathcal{S}$ and selects an action $a \in \mathcal{A}$. The environment deterministically maps the pair $(s,a)$ to a successor state $s'$ by adding the corresponding voxel offset, and returns a scalar reward $r$. The Q-table $Q(s,a)$ is updated according to

\begin{equation} \label{eq:q_learning}
Q_{k+1}(s,a)
= (1-\alpha)\, Q_k(s,a)
+ \alpha \bigl( r + \gamma \max_{a'} Q_k(s',a') \bigr)
\end{equation}

where $\alpha$ is the learning rate and $\gamma$ is the discount factor.  The reward function follows the reproduced baseline: a positive terminal reward at the goal, a large negative reward for collisions or leaving the grid, and a small step penalty to favour shorter paths and discourage wandering. 

After training converges, a greedy policy rolls out a sequence of states 
\begin{equation} \label{eq:state_sequence}
s_0,\, s_1,\, \dots,\, s_N, \qquad s_k \in \mathcal{S}
\end{equation}
connecting the start voxel $s_0$ to the goal voxel $s_N$.  
Mapping each state $s_k$ to its corresponding Cartesian location yields a discrete end-effector path
\begin{equation} \label{eq:rl_position}
x_k^{\mathrm{RL}} \in \mathbb{R}^3, \qquad k = 0, \dots, N
\end{equation}
which is the output of the geometry layer before smoothing.

\subsubsection{Normalized 26-Connected Actions with Geometric Tie-Break}
The core difference between the proposed planner and the baselines lies in the action geometry and the tie-break rule applied during the policy rollout.

In the proposed planner, the index-space neighbour directions are
\begin{equation} \label{eq:discrete_action_vector}
\begin{aligned}
v &= (\Delta i, \Delta j, \Delta k), \\
\Delta i, \Delta j, \Delta k &\in \{-1, 0, 1\}, \qquad 
v \neq (0,0,0)
\end{aligned}
\end{equation}
forming the usual 26-connected neighbourhood. Instead of using these directions directly as grid steps (as in the unnormalized 26-connected baseline), each $v$ is mapped to a Cartesian displacement with a fixed physical step length $\delta>0$:
\begin{equation} \label{eq:continuous_action}
a(v) = \delta\,\frac{v}{\|v\|_2} \in \mathbb{R}^3
\end{equation}
The resulting action set is
\begin{equation} \label{eq:action_set}
\mathcal{A} = \{ a_1, \dots, a_{26} \}
\end{equation}
where each $a_\ell$ has Euclidean norm $\|a_\ell\|_2 = \delta$. This makes the planner isotropic in physical space: axial, face-diagonal, and space-diagonal moves produce the same end-effector displacement magnitude. In the experiments, this eliminates the “long diagonal step’’ bias of the unnormalized 26-connected planner and provides a more uniform base step length for the execution layer. Crucially, the Q-learning state transitions still occur between discrete voxel centers, but the physical path $x_k^{\mathrm{RL}}$ is generated using these normalized displacements.

% When evaluating the greedy policy after training, it often occurs that several actions at state $s$ have similar Q-values. If one simply selects arbitrarily among these, the resulting path exhibits unnecessary zig--zag behaviour and frequent changes of direction. To bias the path toward smoother and safer homotopy classes without modifying the Q-update rule, a geometric tie-break is introduced and applied only when there is more than one near-optimal action.
When evaluating the greedy policy after training, multiple actions at a state $s$ may have similar Q-values. Selecting arbitrarily among them can lead to zig–zag paths and frequent direction changes. To avoid this, a geometric tie-break is introduced, applied only when multiple near-optimal actions exist, biasing the policy toward smoother and safer trajectories without modifying the Q-update rule.

Let
\begin{equation} \label{eq:q_max}
Q_{\max}(s) = \max_{a'} Q(s,a')
\end{equation}
be the largest Q-value at state $s$. For a small threshold $\varepsilon_Q>0$, define the set of $\varepsilon_Q$-near-optimal actions
\begin{equation} \label{eq:tie_actions}
\mathcal{A}_{\text{tie}}(s)
=
\left\{
a \in \mathcal{A} : \bigl| Q(s,a) - Q_{\max}(s) \bigr| < \varepsilon_Q
\right\}.
\end{equation}
Let $a_{k-1}$ be the action taken at the previous step, and let $x$ denote the current end-effector position. For each candidate $a \in \mathcal{A}_{\text{tie}}(s)$, compute
\begin{equation} \label{eq:angle_distance}
\begin{aligned}
\theta(a, a_{k-1}) 
&\quad \text{(turning angle)}, \\
d_{\text{obs}}(x, a) 
&\quad \text{(approx.\ look-ahead clearance)} .
\end{aligned}
\end{equation}
A geometric score is then defined as
\begin{equation} \label{eq:geo_cost}
J_{\text{geo}}(a)
=
-\, w_\theta\, \theta(a,a_{k-1})
\;+\;
w_d\, d_{\text{obs}}(x,a)
\end{equation}
with positive weights $w_\theta$ and $w_d$. The selected action is
\begin{equation} \label{eq:optimal_action}
a^\star
=
\begin{cases}
\displaystyle
\arg\max_{a \in \mathcal{A}_{\text{tie}}(s)} J_{\text{geo}}(a), 
& \text{if } |\mathcal{A}_{\text{tie}}(s)| > 1,\\[6pt]
\displaystyle
\arg\max_{a \in \mathcal{A}} Q(s,a),
& \text{otherwise}.
\end{cases}
\end{equation}
In this way, standard greedy selection $\arg\max_a Q(s,a)$ is recovered whenever there is a unique $\varepsilon_Q$-near-optimal action, while the geometric tie-break only operates when multiple actions have similar Q-values.

For completeness, two baseline planners are also evaluated: \emph{Original}, using a 6-connected action set with grid-resolution step length; and \emph{NoNorm}, using the raw 26-connected index-space directions $v$ as unnormalized physical steps without a tie-break rule.

\subsubsection{Task-space spline smoothing and micro-step control}
The Q-learning planner produces a polyline $\{x_k^{\text{RL}}\}$ with constant step length $\delta$. Although the normalization and tie-breaking mechanisms reduce oscillatory patterns, the path still contains corners. Directly calling inverse kinematics at these points can cause large joint jumps at high-curvature locations. 

To obtain a continuous and differentiable reference trajectory for the execution layer, a two-stage smoothing procedure is applied.

First, three independent cubic splines are fitted to the Cartesian coordinates ($x, y, z$) of $\{x_k^{\text{RL}}\}$, parameterized by the waypoint index $k$. This yields a continuous parametric curve:
\begin{equation} \label{eq:spline_traj}
x^{\text{spl}}(t) \in \mathbb{R}^3, \quad t \in [0, 1]
\end{equation}
Second, the total length $L$ of the curve is approximated, and a desired macro step size $s_{\text{macro}}$ (approximately $5~\text{mm}$ in the implementation) is chosen. The number of macro waypoints is then set as

\begin{equation} \label{eq:macro_steps}
M = \left\lceil \frac{L}{s_{\text{macro}}} \right\rceil
\end{equation}
and parameters $t_i$ are sampled uniformly on $[0,1]$ to define
\begin{equation} \label{eq:ref_points}
x_i^{\text{ref}} = x^{\text{spl}}(t_i), \quad i = 0, \dots, M
\end{equation}
Consecutive macro waypoints $\{ x_i^{\text{ref}} \}$ are thus approximately arc-length uniform.

Finally, a \textit{micro-step constraint} is enforced within the inverse kinematics routine. If the Euclidean distance between consecutive macro-waypoints exceeds a predefined threshold $d_{\max}$ (typically $< 1\,\text{mm}$), intermediate points are inserted via linear interpolation such that every micro-segment satisfies:
\begin{equation} \label{eq:micro_step_constraint}
\|x_{j+1}^{\text{micro}} - x_{j}^{\text{micro}}\|_2 \le d_{\max}
\end{equation}
This constraint guarantees that the end-effector displacement per TP-DLS iteration remains bounded. This stability is corroborated by the convergence results, where the position error per iteration consistently remains well below the $1\,\text{mm}$ threshold.

\subsection{Physics Layer: Task-Priority Damped Least-Squares Execution}

Given the smoothed reference sequence $\{x_i^{\text{ref}}\}_{i=0}^M$, the physics layer computes a joint trajectory $\{q_i\}_{i=0}^M$ on the 7-DoF manipulator. For each reference point $x_i^{\text{ref}}$, the solver starts from the previous configuration $q$ and performs multiple internal iterations of task-priority damped least-squares until the position error falls below a specified tolerance or the maximum iteration count is reached.

\subsubsection{Primary Task: Damped Least-Squares Position Control}

At a given internal iteration, the position error is
\begin{equation} \label{eq:position_error2}
e_p = x_i^{\text{ref}} - p(q) \in \mathbb{R}^3
\end{equation}

The translational Jacobian is
\begin{equation} \label{eq:jacobian_position2}
J_p(q) = J_{1:3,:}(q) \in \mathbb{R}^{3 \times 7}
\end{equation}
and the damped least-squares pseudo-inverse is
\begin{equation} \label{eq:jacobian_pseudoinverse2}
J_p^\dagger(q) = J_p(q)^\top \bigl( J_p(q) J_p(q)^\top + \lambda^2 I_3 \bigr)^{-1}
\end{equation}
with scalar damping factor $\lambda > 0$. Using a scalar position gain $k_p > 0$, the primary joint update is
\begin{equation} \label{eq:delta_q1_2}
\Delta q_1 = J_p^\dagger(q)\, k_p\, e_p
\end{equation}
The associated null-space projector is
\begin{equation} \label{eq:null_space_matrix2}
N_1(q) = I_7 - J_p^\dagger(q) J_p(q)
\end{equation}
Orientation tracking is not enforced in the present experiments: the rotational component of the Jacobian is ignored (or equivalently, its gain is set to zero), allowing the end-effector orientation to adapt freely as the solver exploits redundancy.

\subsubsection{Null-space joint-centering objective}

To avoid configurations near joint limits and maintain a healthy joint-limit margin, a joint-centering objective is added in the null space of the primary task. Let $q_{\min}, q_{\max} \in \mathbb{R}^7$ denote the joint limits, and define
\begin{equation} \label{eq:q_mid}
q_{\text{mid}} = \frac{q_{\min} + q_{\max}}{2}
\end{equation}
as their midpoint. The joint-centering error is
\begin{equation} \label{eq:joint_error}
e_c = q_{\text{mid}} - q
\end{equation}
A simple gradient projection in the null space is applied:
\begin{equation} \label{eq:delta_q2}
\Delta q_2 = \alpha_c \, N_1(q)^\top e_c
\end{equation}
with a small gain $\alpha_c > 0$. Since $\Delta q_2$ lies in the null space of the primary task, it does not affect the end-effector position to first order but redistributes the internal posture. This design results in a more balanced minimum joint-limit margin (JointMargin) for TP-DLS compared with the numerical IK baseline.

The unconstrained joint update is therefore
\begin{equation} \label{eq:delta_q_total}
\Delta q = \Delta q_1 + \Delta q_2
\end{equation}
\subsubsection{Step-Length and Velocity Clipping}

Before applying $\Delta q$ to update the configuration, both a global step-length bound and per-joint bounds are imposed to keep joint displacements and implied velocities within reasonable limits.

Let $\Delta q_{\max} > 0$ denote the scalar step-length bound. If
\begin{equation} \label{eq:delta_q_norm_check}
\|\Delta q\|_2 > \Delta q_{\max}
\end{equation}
the update is scaled as
\begin{equation} \label{eq:delta_q_norm_clip}
\Delta q \leftarrow \Delta q \, \frac{\Delta q_{\max}}{\|\Delta q\|_2}
\end{equation}
Then, for each joint $j = 1,\dots,7$, an individual bound $\Delta q_j^{\max} > 0$ is enforced:
\begin{equation} \label{eq:delta_q_component_clip}
\Delta q_j \leftarrow \operatorname{clip}\bigl( \Delta q_j, -\Delta q_j^{\max}, \Delta q_j^{\max} \bigr)
\end{equation}
where $\operatorname{clip}(\cdot)$ saturates its argument to the specified interval. Given a nominal time step $\Delta t$, these bounds correspond to maximum joint velocities $|\Delta q_j| / \Delta t \le \dot{q}_j^{\max}$.

Finally, the configuration is updated as
\begin{equation} \label{eq:q_update}
q \leftarrow q + \Delta q
\end{equation}
For each reference waypoint $x_i^{\text{ref}}$, this procedure is iterated until $\|e_p\| \le \varepsilon_p$ with $\varepsilon_p = 1~\text{mm}$, or until a maximum number of internal iterations is reached. If any waypoint fails to meet the tolerance within the iteration limit, the entire trajectory is marked as a failure for that solver. For TP-DLS, the recorded per-iteration errors over all waypoints and tasks form the convergence profile, where virtually all iterations remain within the $10^{-4} \sim 10^{-3}~\text{m}$ band, well below the threshold.

As a baseline, a Num-IK executor is also implemented using a conventional numerical inverse kinematics solver. For each reference waypoint, Num-IK solves a full-pose IK problem with the previous solution as initial guess, without task priority, null-space joint centering, or explicit step-length limits. Both executors track the same smoothed reference paths generated by the Improved planner, enabling a fair comparison of joint-space smoothness and safety margins.

The detailed pseudocode of the improved planner and the TP-DLS execution layer is provided in Appendix~A.

\section{Experimental Setup: From Simulation to Data}
This section details the complete experimental workflow conducted on a MATLAB 7-DOF robotic arm simulation platform,ranging from 3D simulation and task set construction to metric evaluation. This manipulator is modeled with a maximum reach of 0.85 meters and operates within a $2.0 \times 2.0 \times 2.0$ meter cubic workspace containing the voxelized grid. Instead of focusing on qualitative visual demonstrations, the objective is to establish a systematic and reproducible framework for rigorously assessing the proposed bridging layer.

% The controlled system is a 7-DoF robotic manipulator with a typical collaborative robot topology, with joint limits set to the same magnitudes as the physical hardware. This work focuses on the safety distance between the end-effector and obstacles. A collision is considered to occur when the clearance between the end-effector and any obstacle becomes non-positive, i.e., when their inter-center distance is no greater than the sum of their radii. Self-collisions are outside the scope of this paper.

The controlled system is a 7-DoF robotic manipulator with a typical collaborative robot topology, with joint limits set to the same magnitudes as the physical hardware. In terms of safety assessment, this work considers not only the clearance between the end-effector and obstacles but also incorporates self-collision detection into the evaluation. A collision is defined to occur when the clearance between the end-effector and any obstacle, or between the robot's own links, becomes non-positive. Specifically, a collision is triggered whenever the inter-center distance between any two bodies is no greater than the sum of their encompassing radii.

\subsection{Scenario Classes and Task Set}

Static environments are grouped into three density classes according to the number of spherical obstacles:
\begin{itemize}
    \item \textbf{Sparse}: 0--30 obstacles.
    \item \textbf{Medium}: 31--99 obstacles.
    \item \textbf{Dense}: at least 100 obstacles.
\end{itemize}

Given a density class and a random seed, the scenario generator deterministically samples obstacle positions and radii. For each scenario, a collision-free initial configuration $q_{\text{start}}$ with sufficient joint margin is selected, its forward kinematics define the start pose, and a target end-effector pose $x_{\text{goal}}$ is specified. Candidate $q_{\text{goal}}$ configurations are obtained by inverse kinematics; solutions in collision or too close to joint limits are discarded and one feasible $q_{\text{goal}}$ is kept.

For every density class, 1,000 start-goal pairs are generated, giving 3,000 tasks in total. All statistics are computed over this task set and reported per density class.

\subsection{Planner Configurations}

At the planning layer, a Q-learning voxel-grid planner is used as a black box. The state space, reward design, and learning hyperparameters are identical in all comparisons; only the action geometry differs:

\begin{itemize}
    \item \textit{\textbf{Original}}: 6-neighbour actions, allowing motion along the three coordinate axes with step size equal to the grid resolution.
    \item \textit{\textbf{NoNorm}}: unnormalized 26-neighbour actions, allowing transitions to all adjacent voxels with different Euclidean step lengths for axial and diagonal moves.
    \item \textit{\textbf{Improved (proposed)}}: 26-neighbour actions normalized to a uniform physical step length, combined with a geometric tie-breaking rule based on turning angle and local clearance when Q-values are nearly equal.
\end{itemize}

All planners share the same learning rate, discount factor, and $\varepsilon$-greedy exploration. After convergence, each planner generates a discrete end-effector path for each task. The voxel sequences are mapped to Cartesian space, fitted with cubic splines, and resampled at approximately equal arc length to obtain a smooth reference path. This path is used to evaluate planner geometry (length, turns, clearance) and as a common input to the execution layer.

\subsection{Execution Layer Configurations}

To isolate execution effects, all executors operate on the same reference paths from the Improved planner; only the inverse kinematics method differs.

\textbf{Num-IK.} A toolbox-based numerical IK baseline. For each reference waypoint, inverseKinematics is called independently, using the previous solution as the initial guess. Redundancy is not exploited and joint increments are not explicitly bounded.

\textbf{TP-DLS (proposed).} A task-priority damped least-squares scheme with end-effector tracking as the primary task and joint-centering in the null space. Joint increments are norm- and component-clamped at each step to regulate velocity and acceleration.

In this way, differences in execution quality can be attributed to the IK strategy rather than to differences in the planned path.

\subsection{Data Pipeline and Metrics}

For each task, planner outputs and joint trajectories are processed by a unified script to compute all metrics.

Planner-level metrics include success rate, path length, number of turns, step count, and minimum end-effector–obstacle distance. Execution-level metrics include the 95th percentile of joint increment norm, peak joint velocity and acceleration, minimum joint-limit margin, number of retries, and total computation time.

For planner comparison, failed tasks are assigned fixed penalty values for length, turns, and steps to jointly reflect feasibility and path quality. For execution-layer comparison, the planner is fixed to the improved variant.

\section{Results and Discussion}

\begin{figure}[htbp]
    \centering
    \includegraphics[width=\columnwidth]{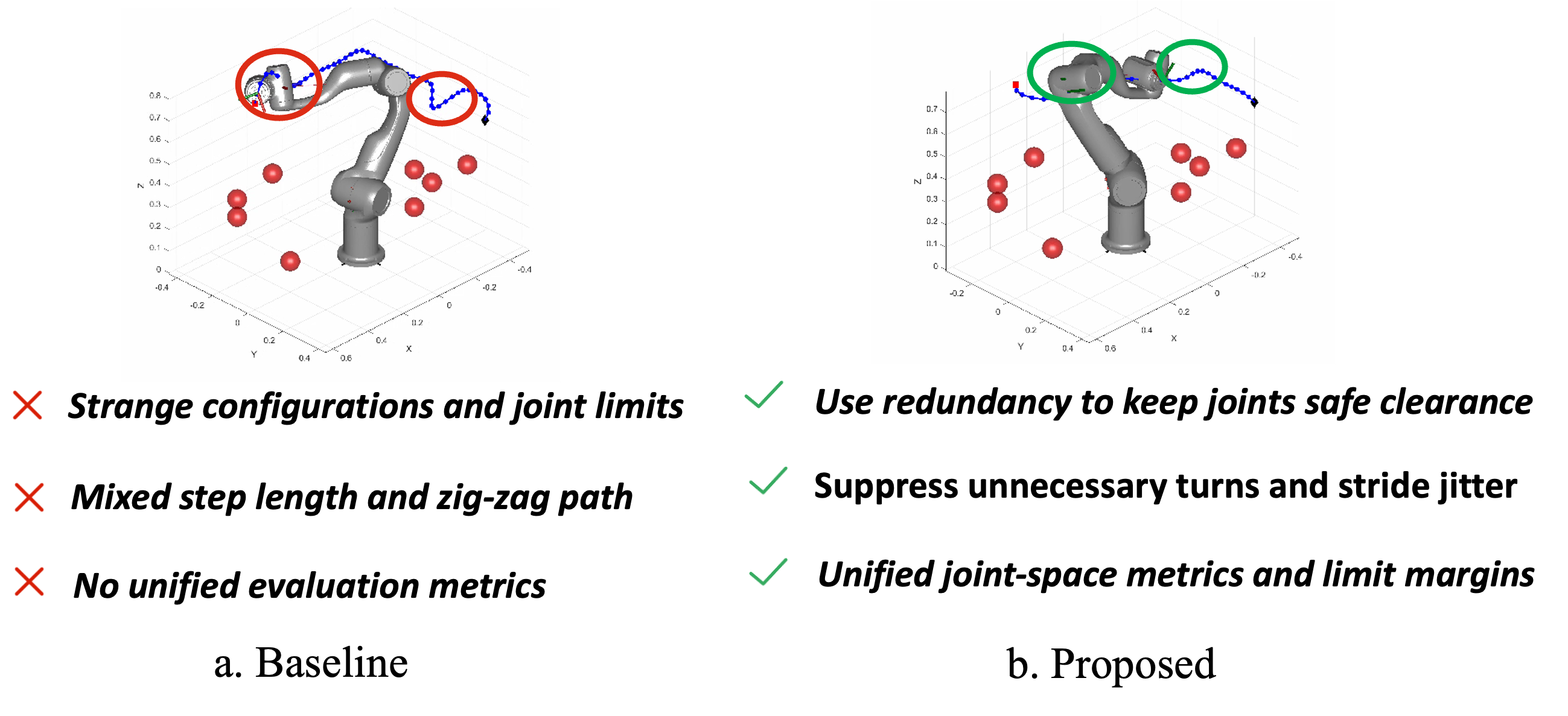}
    \caption{\textbf{Qualitative comparison on a 7-DoF arm.} (a) Baseline (unnormalised actions + Num-IK) shows irregular steps and joint-limit
configurations.
(b) Proposed bridge (normalised 26-neighbour + TP–DLS) equalises step size and
keeps joints away from limits.}
    \label{fig:result}
\end{figure}

Fig\ref{fig:result} provides a qualitative comparison demonstrating how the proposed bridge eliminates irregular steps and maintains safer joint configurations compared to the baseline approach.

\subsection{Planning Interface Regularization}
Table~\ref{tb:planner} reports the planning metrics for the three planners over 150 queries. The \textit{Original} and \textit{NoNorm} baselines reach only 0.56–0.70 success rate, with most failures occurring in dense scenes. The proposed \textit{Improved} planner achieves 1.00 success rate under all obstacle densities.

Regarding geometric quality, the \textit{Improved} method consistently produces shorter paths with fewer turns than both baselines, especially in dense environments. Minimum-clearance values, however, remain nearly identical across all methods, indicating that the improvements do not trade clearance for path length. Instead, step normalization and geometric tie-breaking reduce the irregular step alternation and backtracking inherent in the unnormalized 26-neighbor expansion. Step-size statistics further show that the \textit{Improved} planner yields highly uniform waypoint spacing, which is beneficial for subsequent spline interpolation and DLS tracking.

\begin{table*}[t] % 注意这里加了星号，位置改为 [t] 顶部
\begin{center}
\caption{Three Planners Under Different Obstacle Densities}\label{tb:planner}
\begin{tabular}{ccccccc}
\hline 
Density & Algorithm & Success rate & Length & Turns & MinClearance & Steps \\\hline
Sparse & Original & 0.68 & 192.572 & 1862.9 & 0.213 & 2126.72 \\
Sparse & NoNorm & 0.6 & 283.35 & 1451.42 & 0.213 & 2563.84 \\
Sparse & Improved & \textbf{1.0} & \textbf{0.602} & \textbf{4.26} & 0.216 & \textbf{7.02} \\
Medium & Original & 0.68 & 168.516 & 1681.98 & 0.157 & 1926.16 \\ 
Medium & NoNorm & 0.7 & 195.7 & 1122.5 & 0.152 & 1725.12 \\
Medium & Improved & \textbf{1.0} & \textbf{0.598} & \textbf{3.98} & 0.165 & \textbf{6.98} \\
Dense & Original & 0.56 & 396.468 & 3901.6 & 0.125 & 3965.68 \\
Dense & NoNorm & 0.58 & 499.377 & 3237.8 & 0.119 & 3783.9 \\
Dense & Improved & \textbf{1.0} & \textbf{0.68} & \textbf{5.02} & 0.138 & \textbf{7.8} \\
\hline
\end{tabular}
\end{center}
\end{table*}

% \begin{table*}[t]
% \begin{center}
% \caption{Two Execution Layers Under the Same Reference Trajectory}\label{tb:execution_layers}
% \begin{tabular}{ccccccccc}
% \hline
% Density & Solver & Success\_rate & DQ\_95 & MaxVel & MaxAcc & JointMargin & Backtracks & Time \\ \hline
% Sparse & Num-IK & 1.0 & 0.8211 & 141.8634 & 12701.6607 & 0.0053 & 7.18 & 0.7231 \\
% Sparse & TP-DLS & 1.0 & \textbf{0.0113} & \textbf{84.2102} & \textbf{3943.5816} & \textbf{0.0126} & 19.44 & 35.9558 \\ 
% Medium & Num-IK & 1.0 & 0.6415 & 190.3101 & 16793.8613 & 0.0045 & 6.5 & 0.6859 \\
% Medium & TP-DLS & 1.0 & \textbf{0.0113} & \textbf{79.747} & \textbf{3860.0759} & \textbf{0.0112} & 37.6 & 36.7482 \\
% Dense & Num-IK & 1.0 & 0.937 & 175.8501 & 15659.236 & 0.0002 & 5.48 & 0.6497 \\
% Dense & TP-DLS & 1.0 & \textbf{0.0088} & \textbf{91.651} & \textbf{4438.005} & \textbf{0.014} & 22.16 & 29.0056 \\
% \hline
% \end{tabular}
% \end{center}
% \end{table*}

\begin{table*}[t]
\begin{center}
\caption{Two Execution Layers Under the Same Reference Trajectory}\label{tb:execution_layers}
\begin{tabular}{ccccccccc}
\hline
Density & Solver & Success\_rate & DQ\_95 & MaxVel & MaxAcc & JointMargin & Backtracks  \\ \hline
Sparse & Num-IK & 1.0 & 0.8211 & 141.8634 & 12701.6607 & 0.0053 & 7.18 & 0.7231 \\
Sparse & TP-DLS & 1.0 & \textbf{0.0113} & \textbf{84.2102} & \textbf{3943.5816} & \textbf{0.0126} & 19.44  \\ 
Medium & Num-IK & 1.0 & 0.6415 & 190.3101 & 16793.8613 & 0.0045 & 6.5  \\
Medium & TP-DLS & 1.0 & \textbf{0.0113} & \textbf{79.747} & \textbf{3860.0759} & \textbf{0.0112} & 37.6  \\
Dense & Num-IK & 1.0 & 0.937 & 175.8501 & 15659.236 & 0.0002 & 5.48 \\
Dense & TP-DLS & 1.0 & \textbf{0.0088} & \textbf{91.651} & \textbf{4438.005} & \textbf{0.014} & 22.16 \\
\hline
\end{tabular}
\end{center}
\end{table*}

\subsection{Impact of Execution Layer on Trajectory Quality}
For the execution layer comparison, the planner is fixed to \textit{Improved} for all 150 tasks. Since both solvers track identical task-space trajectories, metrics such as path length, number of turns, and minimum clearance remain virtually unchanged, and both maintain position errors within 1~mm for all reference points, achieving a success rate of approximately 100\%. 

Table~\ref{tb:execution_layers} reveals that TP-DLS exhibits an order-of-magnitude advantage over the numerical IK baseline in terms of $DQ_{95}$. Across all obstacle densities, TP-DLS keeps $DQ_{95}$ on the order of $10^{-2}$~rad, whereas \textit{NumIK} ranges from $10^{-1}$ to $10^{0}$~rad. In representative trajectories, maximum joint increments remain below the 0.1-rad warning threshold, causing the curve to remain nearly flat. Likewise, TP-DLS substantially reduces $MaxVel$ and $MaxAcc$, suppressing the abrupt joint transitions and dynamic spikes typical of numerical IK.

% Statistics for $JointMargin$ show that both solvers maintain similar distances from joint limits, with TP-DLS exhibiting a slight advantage due to its null-space joint-centering. This improvement comes at a higher computational cost: $Backtracks$ and $Time$ are noticeably greater than for \textit{NumIK}, reflecting the use of small-step iterations and retries in locally challenging regions. As the framework operates offline, this overhead is acceptable.

Statistics for $JointMargin$ show that both solvers keep comparable distances from joint limits, with TP-DLS enjoying a slight benefit from null-space joint-centering. This smoothness comes at a clear computational cost: $Backtracks$ and $Time$ are substantially higher than for \textit{NumIK}. TP-DLS deliberately advances in thousands of very small increments and repeatedly resolves IK in locally difficult regions to achieve extremely smooth joint evolution. As the framework is explicitly designed for offline pre-planning, the increased computation time is an intentional and acceptable trade-off for motion quality.

In representative dense scenarios, the final position error of interpolated reference points for TP-DLS remains within $10^{-4}$ to $10^{-3}$~m, consistently below the 1~mm threshold. Most errors cluster near $10^{-4}$~m, with only a few larger values during the initial approach phase, indicating stable convergence and reliable trajectory tracking across tasks.

\section{Conclusion}
% This work addresses the common robotic RL planning paradigm characterized by voxel grids + finite neighborhood actions + Q-learning. Without modifying the underlying planning algorithm, we propose a bridging framework between discrete planning and continuous execution for 7-DoF redundant manipulators. On the planning side, the introduction of step-normalized 26-neighbor actions and a geometric tie-breaking mechanism explicitly eliminates the geometric anisotropy of discrete actions. This significantly reduces path length and the number of turns while improving success rates in complex obstacle scenarios, all without compromising minimum clearance. On the execution side, a TP-DLS inverse kinematics layer, combined with task-space spline smoothing and step-limited interpolation, to convert the discrete path into a smooth joint trajectory in an offline manner, achieving millimeter-level end-effector accuracy and substantially improved joint smoothness. Consequently, the 95th-percentile joint increments and dynamic peaks are compressed to levels far lower than those of traditional numerical IK.
This work addresses the common robotic RL planning paradigm of voxel grids with finite neighborhood actions and Q-learning by proposing a bridging framework between discrete planning and continuous execution for 7-DoF redundant manipulators, without modifying the planner itself. On the planning side, step-normalized 26-neighbor actions with a geometric tie-breaking rule regularize action geometry, reducing path length and unnecessary turns while preserving clearance and improving success rates in complex scenes. On the execution side, a standard TP-DLS inverse kinematics method, combined with task-space spline smoothing and step-limited interpolation, converts the discrete path offline into a smooth joint trajectory with millimeter-level end-effector accuracy and markedly improved joint smoothness, greatly reducing high-percentile joint increments and dynamic peaks compared to traditional numerical IK.

Since this bridging layer relies solely on the discrete end-effector path output by the planner and remains agnostic to the specific RL algorithm or network structure, it serves as a generic interface that effectively decouples existing grid-based RL planning from the continuous control of redundant manipulators. Future work will integrate this method with safety filtering techniques to achieve online execution with formal safety guarantees. Additionally, we aim to validate its applicability and limitations on physical robot platforms and in more complex tasks, such as multi-arm coordination and dexterous manipulation.

\bibliography{ifacconf}             % bib file to produce the bibliography
                                                     % with bibtex (preferred)
                                                   
%\begin{thebibliography}{xx}  % you can also add the bibliography by hand

%\bibitem[Able(1956)]{Abl:56}
%B.C. Able.
%\newblock Nucleic acid content of microscope.
%\newblock \emph{Nature}, 135:\penalty0 7--9, 1956.

%\bibitem[Able et~al.(1954)Able, Tagg, and Rush]{AbTaRu:54}
%B.C. Able, R.A. Tagg, and M.~Rush.
%\newblock Enzyme-catalyzed cellular transanimations.
%\newblock In A.F. Round, editor, \emph{Advances in Enzymology}, volume~2, pages
%  125--247. Academic Press, New York, 3rd edition, 1954.

%\bibitem[Keohane(1958)]{Keo:58}
%R.~Keohane.
%\newblock \emph{Power and Interdependence: World Politics in Transitions}.
%\newblock Little, Brown \& Co., Boston, 1958.

%\bibitem[Powers(1985)]{Pow:85}
%T.~Powers.
%\newblock Is there a way out?
%\newblock \emph{Harpers}, pages 35--47, June 1985.

%\bibitem[Soukhanov(1992)]{Heritage:92}
%A.~H. Soukhanov, editor.
%\newblock \emph{{The American Heritage. Dictionary of the American Language}}.
%\newblock Houghton Mifflin Company, 1992.

%\end{thebibliography}

\appendix
\section{Pseudocode of the Planning–Execution Bridge}    % Each appendix must have a short title.
\begin{algorithm}[H]
  \caption{Improved RL planner and reference path}
  \label{alg:planner}
  \begin{algorithmic}[1]
    \Require start/goal $x_s,x_g\in\mathbb{R}^3$; obstacles $\mathcal{O}$;
            Q-table $Q$; step $\delta$; macro step $s_{\text{mac}}$;
            micro step $d_{\max}$
    \Ensure  reference waypoints $X_{\text{ref}} = \{x_i\}$

    \State $s \gets \textsc{VoxelFromPos}(x_s)$
    \State $s_G \gets \textsc{VoxelFromPos}(x_g)$
    \State $k \gets 0$;\quad
           $X_{\text{RL}} \gets [\,\textsc{PosFromVoxel}(s)\,]$;\quad
           $a_{\text{prev}} \gets \textsc{Null}$

    \While{$s \neq s_G$ \textbf{and} $k < K_{\max}$} \Comment{RL rollout on normalized 26-neighbourhood}
      \State $Q_{\max} \gets \max_a Q(s,a)$
      \State $\mathcal{A}_{\text{tie}} \gets
              \{ a\in\mathcal{A} \mid |Q(s,a)-Q_{\max}| < \varepsilon_Q \}$
      \If{$|\mathcal{A}_{\text{tie}}| > 1$} \Comment{true tie: several near-optimal actions}
        \State $a^\star \gets
          \arg\max_{a\in\mathcal{A}_{\text{tie}}}
          J_{\text{geo}}(a,a_{\text{prev}},X_{\text{RL}}[k],\mathcal{O})$
      \Else \Comment{unique winner: fall back to greedy}
        \State $a^\star \gets \arg\max_{a\in\mathcal{A}} Q(s,a)$
      \EndIf

      \State $s \gets \textsc{NextState}(s,a^\star,\delta)$
      \State $a_{\text{prev}} \gets a^\star$
      \State append $\textsc{PosFromVoxel}(s)$ to $X_{\text{RL}}$
      \If{$\textsc{Collision}(X_{\text{RL}}[k+1],\mathcal{O})$}
        \State \textbf{break}
      \EndIf
      \State $k \gets k+1$
    \EndWhile

    \State $X^{\text{mac}} \gets
            \textsc{SplineResample}(X_{\text{RL}},s_{\text{mac}})$
    \State $X_{\text{ref}} \gets
            \textsc{Subdivide}(X^{\text{mac}},d_{\max})$
  \end{algorithmic}
\end{algorithm}
\begin{algorithm}[H]
  \caption{TP--DLS execution along the reference path}
  \label{alg:tpdls}
  \begin{algorithmic}[1]
    \Require $X_{\text{ref}}=\{x_i^{\text{ref}}\}$; robot model; end-effector name;
             $q_0$; joint limits $q_{\min},q_{\max}$;
             damping $\lambda$; gain $k_p$; null-space gain $\alpha_c$;
             bounds $\Delta q_{\max},\{\Delta q_{\max,j}\}$;
             tolerance $\varepsilon_p$; max iterations $K_{\max}$
    \Ensure  joint trajectory $Q=\{q[i]\}$; success flag

    \State $q \gets q_0$;\quad
           $q_{\text{mid}} \gets \tfrac{1}{2}(q_{\min}+q_{\max})$;\quad
           $Q[0] \gets q$

    \For{$i = 0$ \textbf{to} $|X_{\text{ref}}|-1$}
      \State $x^\star \gets x_i^{\text{ref}}$;\quad $k \gets 0$
      \Repeat
        \State $x   \gets \textsc{ForwardPos}(\text{robot},q)$
        \State $e_p \gets x^\star - x$
        \If{$\lVert e_p\rVert \le \varepsilon_p$}
          \State \textbf{break}
        \EndIf

        \State $J   \gets \textsc{GeomJacobian}(\text{robot},q)$
        \State $J_p \gets J_{1:3,:}$ \Comment{translational block}
        \State $J_p^\dagger \gets J_p^\top (J_p J_p^\top + \lambda^2 I_3)^{-1}$
        \State $\Delta q_1 \gets J_p^\dagger (k_p e_p)$

        \State $N_1 \gets I_7 - J_p^\dagger J_p$
        \State $e_c \gets q_{\text{mid}} - q$
        \State $\Delta q_2 \gets \alpha_c N_1^\top e_c$

        \State $\Delta q  \gets \Delta q_1 + \Delta q_2$
        \State $\Delta q  \gets \textsc{ClipNorm}(\Delta q,\Delta q_{\max})$
        \For{$j = 1$ \textbf{to} $7$}
          \State $\Delta q_j \gets
                 \textsc{Clip}(\Delta q_j,-\Delta q_{\max,j},\Delta q_{\max,j})$
        \EndFor

        \State $q \gets q + \Delta q$;\quad $k \gets k+1$
      \Until{$k \ge K_{\max}$}

      \If{$\lVert e_p\rVert > \varepsilon_p$}
        \State \textbf{return} $(Q,\textsc{false})$
      \EndIf
      \State $Q[i+1] \gets q$
    \EndFor
    \State \textbf{return} $(Q,\textsc{true})$
  \end{algorithmic}
\end{algorithm}

\end{document}